\documentclass[sigconf,prologue,dvipsnames]{acmart}

\AtBeginDocument{%
  \providecommand\BibTeX{{%
    \normalfont B\kern-0.5em{\scshape i\kern-0.25em b}\kern-0.8em\TeX}}}

\copyrightyear{2024}
\acmYear{2024}
\setcopyright{rightsretained}
\acmConference[HRI '24]{Proceedings of the 2024 ACM/IEEE International Conference on Human-Robot Interaction}{March 11--14, 2024}{Boulder, CO, USA}
\acmBooktitle{Proceedings of the 2024 ACM/IEEE International Conference on Human-Robot Interaction (HRI '24), March 11--14, 2024, Boulder, CO, USA}
\acmDOI{10.1145/3610977.3634987}
\acmISBN{979-8-4007-0322-5/24/03}

\makeatletter
\gdef\@copyrightpermission{
  \begin{minipage}{0.3\columnwidth}
   \href{https://creativecommons.org/licenses/by/4.0/}{\includegraphics[width=0.90\textwidth]{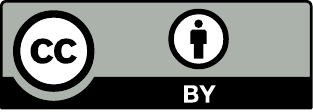}}
  \end{minipage}\hfill
  \begin{minipage}{0.7\columnwidth}
   \href{https://creativecommons.org/licenses/by/4.0/}{This work is licensed under a Creative Commons Attribution International 4.0 License.}
  \end{minipage}
  \vspace{5pt}
}
\makeatother

\usepackage{xcolor}
\usepackage{graphicx}
\usepackage{subcaption}
\usepackage{amsmath}

\usepackage{balance}

\usepackage{color}

\newcommand{\rob}{R}
\newcommand{\hum}{H}
\newcommand{\hor}{T}
\newcommand{\st}{\ensuremath{s}}
\newcommand{\sthis}{\ensuremath{\textbf{s}}}
\newcommand{\ac}{\ensuremath{a}}
\newcommand{\ob}{\ensuremath{o}}
\newcommand{\obhis}{\ensuremath{\textbf{o}}}
\newcommand{\St}{\ensuremath{\mathcal{S}}}
\newcommand{\Ac}{\ensuremath{\mathcal{A}}}
\newcommand{\Ob}{\ensuremath{\mathcal{O}}}

\newcommand{\reward}{r}
\newcommand{\policy}{\ensuremath{\pi}}
\newcommand{\transf}{\ensuremath{f}}
\newcommand{\transmat}{\ensuremath{F}}
\newcommand{\weight}{\ensuremath{\theta}}

\newcommand{\rep}{\ensuremath{\phi}}
\newcommand{\repH}{\ensuremath{\rep_H(\obhis_H)}}
\newcommand{\repR}{\ensuremath{\rep_R(\obhis_R)}}
\newcommand{\repspace}{\ensuremath{\Phi}}
\newcommand{\repfuncH}{\ensuremath{\phi_H}}
\newcommand{\repfuncR}{\ensuremath{\phi_R}}
\newcommand{\simfunc}{\ensuremath{\psi}}

\newcommand\figref[1]{Fig.~\ref{#1}}
\newcommand\equref[1]{Eq.~\eqref{#1}}
\newcommand\secref[1]{Sec.~\ref{#1}}

\begin{document}

\title{Aligning Human and Robot Representations}

\author{Andreea Bobu}
\authornote{Equal contribution. This research is supported by the Air Force Office of Scientific Research (AFOSR), NSF HCC, NSF Graduate Research Fellowship, and Open Philanthropy.}
\email{abobu@berkeley.edu}
\affiliation{\institution{University of California, Berkeley} \country{United States of America}}

\author{Andi Peng}
\authornotemark[1]
\email{andipeng@mit.edu}
\affiliation{\institution{MIT} \country{United States of America}}

\author{Pulkit Agrawal}
\email{pulkitag@mit.edu}
\affiliation{\institution{MIT} \country{United States of America}}

\author{Julie A. Shah}
\email{julie_a_shah@csail.mit.edu}
\affiliation{\institution{MIT} \country{United States of America}}

\author{Anca D. Dragan}
\email{anca@berkeley.edu}
\affiliation{\institution{University of California, Berkeley} \country{United States of America}}

\renewcommand{\shortauthors}{Andreea Bobu, Andi Peng, Pulkit Agrawal, Julie A. Shah, \& Anca D. Dragan}

\begin{abstract}
  To act in the world, robots rely on a \textit{representation} of salient task aspects: for example, to carry a coffee mug, a robot may consider movement efficiency or mug orientation in its behaviour. However, if we want robots to act \textit{for and with people}, their representations must not be just functional but also reflective of what humans care about, i.e. they must be \textit{aligned}. We observe that current learning approaches suffer from \textit{representation misalignment}, where the robot's learned representation does not capture the human's representation. We suggest that because humans are the ultimate evaluator of robot performance, we must \textit{explicitly} focus our efforts on aligning learned representations with humans, \textit{in addition to} learning the downstream task. We advocate that current representation learning approaches in robotics should be studied from the perspective of how well they accomplish the objective of representation alignment. We mathematically define the problem, identify its key desiderata, and situate current methods within this formalism. We conclude by suggesting future directions for exploring open challenges.
\end{abstract}


\maketitle

\section{Introduction}
\label{sec:intro}

\begin{figure}
\centering
\includegraphics[width=0.47\textwidth]{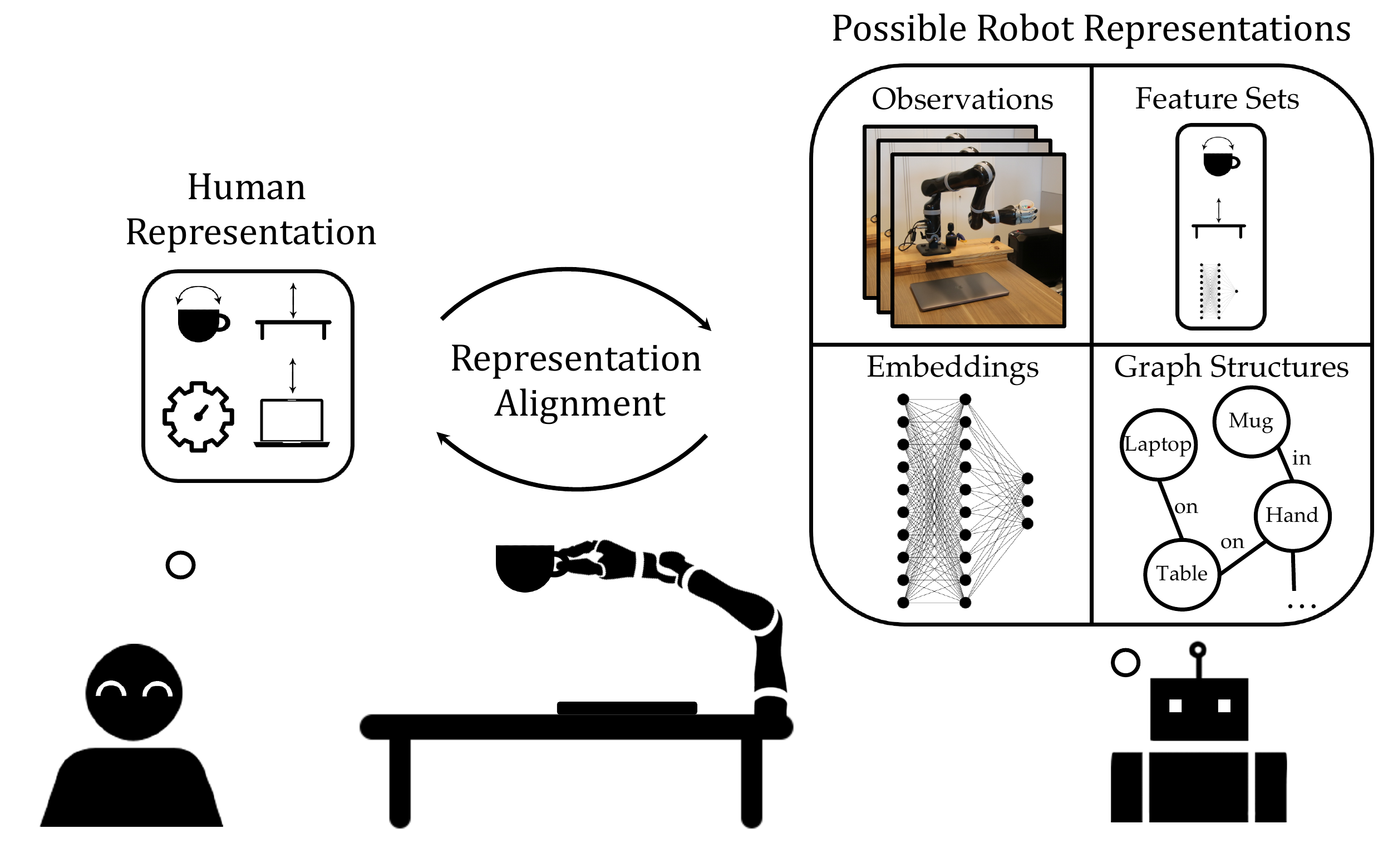}\\
\caption{We formalize representation alignment as the search for a robot task representation that is \textit{easily able} to capture the true human task representation. We review four categories of current robot representations and summarize their key takeaways and tradeoffs.}
 \label{fig:frontfig}
\end{figure}

In the HRI community, we aspire to build robots that perform tasks that human users want them to perform. To do so, robots need good \textit{representations} of salient task aspects.
For example, in \figref{fig:frontfig}, to carry a coffee mug, the robot considers efficiency, mug orientation, and distance from the user's possessions in its behaviour.
There are two paradigms for learning representations:
one that \textit{explicitly} builds in structure for learning task aspects, e.g. feature sets or graphs, and one that \textit{implicitly} extracts task aspects by mapping input directly to desired behaviour, e.g. end-to-end approaches~\cite{levine2020offline,ross2011reduction}. While explicit structure is useful for capturing relevant task aspects, it's often impossible to comprehensively define all aspects that may matter to the downstream task; meanwhile, implicit methods circumvent this problem by allowing neural networks to automatically extract representations, but they are prone to capturing \textit{spurious correlations}~\cite{levine2020offline}, resulting in potentially arbitrarily bad robot behaviour under distribution shift between train and test conditions~\cite{paudel2022learning}.

Our observation is that many failures in robot learning, including the ones above, result from a \textit{mismatch between the human's representation and the one learned by the robot}; in other words, their representations are \textit{misaligned}. From this perspective, these failures illuminate that if we truly wish to learn good representations -- if we truly want robots that do what humans want -- we must explicitly focus on the foundational problem: \textit{aligning robot and human representations}. In this paper, we offer a unifying lens for the HRI community to view existing and future solutions to this problem.


We review over 100 papers in the representation learning literature in robotics from this perspective. We first define a unifying mathematical objective for an aligned representation based on four desiderata: value alignment, generalizable task performance, reduced human burden, and explainability. We then conduct an in-depth review of four common representations (\figref{fig:frontfig}): the identity representation, feature sets, feature embeddings, and graphical structures -- illustrating the deltas each falls short in with respect to the desiderata. From situating each representation in our formalism, we arrive at the following key takeaway: a better structured representation affords better alignment and therefore better task performance, but always with the unavoidable tradeoff of more human effort. This effort can be directed in three ways: 1) representations that operate directly on the observation space, e.g. end-to-end methods, direct effort at increasing task data to avoid spurious correlations; 2) representations that build explicit task structure, e.g. graphs or feature sets, direct effort at constructing and expanding the representation; and 3) representations that learn directly from implicit human representations, e.g. self-supervised models, direct effort at creating good proxy tasks. 

Our paper is untraditional in that it is much like a survey paper, except there is little work that directly addresses the representation alignment problem we pose. Instead, we offer a retrospective on works that focus on learning task representations in robotics with respect to our desiderata. Our review provides a unifying lens to think about the current gaps present in the robot learning literature as defined by a common language, or in other words, a roadmap for thinking about challenges present in current and future solutions in a principled way. We conclude by suggesting key open directions.

\section{The Desired Representation}
\label{sec:desiderata}

Before formalizing the problem, we build intuition for the desiderata defining aligned representations.

\smallskip
\noindent\textbf{Value Alignment.} Learning human-aligned representations can aid with \textit{value alignment}~\cite{amodei2016concrete}, enabling robots to perform well under the human's desired objective rather than optimize misspecified objectives that lead to unintended side-effects. In ``reward hacking'' scenarios~\cite{amodei2016concrete}, if the representation of human intents is ill-defined or insufficient, the reward learned on top of it will optimize for the wrong human objective. In the canonical example of a robot tasked with sweeping dust off the floor~\cite{russell2010artificial}, an optimal policy for the reward ``maximize dust collected off the floor'' leads the robot to dump dust just to immediately sweep it up again. In this case, the reward is defined on top of a representation that is \textit{under-specified}, i.e. the amount of dust that is collected, and fails to capture other important features, e.g. covering the whole house, not adding dust on the floor, etc. Explicitly learning a representation aligned with the human's may ensure that the robot \textit{fully} captures the causal task features that make the desired human objective \textit{realizable}.

\smallskip
\noindent\textbf{Generalizable Task Learning}. A human-aligned representation may afford more generalizable task learning~\cite{du2020fewshot,arora2020replearn,peng2023diagnosis}. A central problem in robot learning is capturing diverse behaviors across different environments and user preferences~\cite{levine2020offline,paudel2022learning}.
While domains like natural language or vision have achieved impressive performance across tasks by using large-scale datasets~\cite{ramesh2021zero,brown2020language,radford2021learning}, robot learning is bottlenecked by our ability to collect diverse data that captures the complexity of the world. 
Without it, neural networks may learn non-causal correlates in the input space~\cite{ilyas2019adversarial,dehaan2019causal}. Thus, learning objectives that operate directly on high-dimensional input spaces suffer from spurious correlations, where the implicit representation may contain features that are \textit{irrelevant} to the task~\cite{agrawal2022task}. Consequently, the learned network may be based on these correlated irrelevant features that appear causal in-distribution, but fail under distribution shift. Explicitly aligning robot representations with those used by humans may avoid learning irrelevant features and, thus, may afford more generalizable and robust task learning.

\smallskip
\noindent\textbf{Reducing Human Burden.} Operating on human-aligned representations may reduce teaching burden. In our above two scenarios, where human guidance is either task demonstrations or specified rewards, if we had unlimited human time and effort, we would be able to provide a perfect task representation, i.e. a demonstration of the task in every environment for every user~\cite{eysenbach2018diversity}, or a reward function that specifies every feature any user may find relevant for performing the task in any environment~\cite{hadfield2017inverse}, and then fit the data with an arbitrarily complex function such as a neural network. In practice, both scenarios are infeasible with low sample complexity, and therefore motivate the explicit need for representations that align with humans on the task abstraction level~\cite{abel2021expressivity,ho2019value,abel2018state}.

\smallskip
\noindent\textbf{Explainability.} We want representations that enable system transparency for ethical, legal, safety, or usability reasons~\cite{glanois2021survey,alharin2020reinforcement}. Current methods range from generating post-hoc explanations \cite{glanois2021survey,bewley2021tripletree}, text descriptions of relational MDPs \cite{hayes2017improving,sanner2005simultaneous}, or saliency maps~\cite{greydanus2018visualizing} for explaining behavior. 
However, system interpretability should not only be considered during deployment, but also be embedded within the design process itself~\cite{garcia2015comprehensive,gupta2019explain}. 
Explicitly aligning representations with humans' can create a more streamlined process for ensuring that representations are primed for human understanding~\cite{rudin2019stop}. 

\smallskip
\noindent\textbf{Desideratum 1:} The representation should capture \textit{all} the relevant aspects of the desired task, i.e., the human's true objective should be \textit{realizable} when using the representation for task learning. 

\smallskip
\noindent\textbf{Desideratum 2:} 
The representation should not capture \textit{irrelevant} aspects of the desired task, i.e., the representation should not be based on spurious correlations. 

\smallskip
\noindent\textbf{Desideratum 3:} Human guidance for learning the representation should demand \textit{minimal} time and effort, i.e., the human's representation should be \textit{easily recoverable} from data. 

\smallskip
\noindent\textbf{Desideratum 4:} The representation should enable system \textit{interpretability} and \textit{explainability}, affording safe, transparent systems that can integrate with human users in the real world.

\smallskip 

We henceforth refer to these desiderata as D1-4, mathematically operationalize them in the context of learning robot representations from humans, and situate how prior works relate to these goals.
\section{Problem Formulation}
\label{sec:formulation}




\noindent\textbf{Setup.}
We consider cases where a robot $\rob$ seeks to learn how to perform a task desired by a human $\hum$. The two agents live in state $\st\in\St$ and execute actions $\ac_H\in\Ac_H$ and $\ac_R\in\Ac_R$. 
The robot's goal is to learn a task expressed via a \textit{reward function} $\reward^*: \St \rightarrow \mathbb{R}$ capturing the human's preference over states. The human knows the desired task, and, thus, implicitly knows $\reward^*$ and how to act accordingly via a \textit{policy} $\policy^*(\ac_H \mid \st) \in [0,1]$, but the robot does not 
and has to learn that from the human. 

We consider two popular robot learning approaches: \textit{imitation learning}, where we learn the human's policy for solving the task, and \textit{reward learning}, where we learn the reward function describing the task. The approaches have different trade-offs: imitation learning does not require modeling the human and simply replicates their actions~\cite{osa2018algorithmic,abbeel2004apprenticeship}, but in doing so it also replicates their suboptimality and can't generalize well to changing dynamics or state distributions~\cite{levine2020offline,torabi2018behavioural}; meanwhile, reward learning attempts to capture \textit{why} a specific behaviour is desirable and, thus, can generalize better to novel scenarios~\cite{abbeel2004apprenticeship} but requires assuming a human model and large amounts of data~\cite{fu2018learning,Reddy2020SQILIL}.

\smallskip
\noindent\textbf{Partial Observability and Representations.}
We first examine how state $\st$ should be represented.
In theory, the state $\st$ could comprehensively capture the ``true'' components of the world down to their atomic elements, but in practice such a hypothetical state is neither fully observable nor useful. Instead, we assume that neither agent has the full state but they each \emph{observe} it via observations $\ob_H \in \Ob_H$ and $\ob_R \in \Ob_R$. The robot's observations $\ob_R$ come from its (possibly noisy and non-deterministic) sensors $P(\ob_R \mid s)$, e.g. robot joint angles, RGB-D images, object poses and bounding boxes, etc. The human also senses observations $\ob_H$ via their ``sensors'', e.g. retinal inputs, audio signals, etc., which we could model according to $P(\ob_H \mid s)$. Due to partial observability, both the robot and the human use the \textit{history} of $t$ observations $\obhis_R = (\ob_R^1, ..., \ob_R^t)\in\Ob^t_R$ and $\obhis_H = (\ob_H^1, ..., \ob_H^t)\in\Ob^t_H$, respectively, as a proxy for the state -- or sequence of states -- they observe $\sthis = (\st^1, ..., \st^t)\in\St^t$. We assume that $\obhis_R$ and $\obhis_H$ correspond to the same $\sthis$.

Neuroscience and cognitive psychology literature suggest that humans don't estimate the state directly from the complete $\obhis_H$~\cite{birman2019filter}. Instead, people focus on what's important for their task, often ignoring task-irrelevant attributes~\cite{callaway2021fixation}, and build a task-relevant \textit{representation} to help them solve the task~\cite{bonnen2021vvs}. We, thus, assume that when humans think about how to complete or evaluate a task, they operate on a representation $\repH$ given by the transformation $\repfuncH : \Ob_H^t \rightarrow \repspace_H$, 
which determines which information in $\obhis_H$ to focus on and how to combine it into something useful for the task. For example, to determine if two novel objects have the same shape, a human might first look around both of them (gather a sequence of visual information $\obhis_H$) to build an approximate 3D model (representation $\repH$). Intuitively, we can think of such a representation as an estimate of the task-relevant components of the state, in lieu of the true unknown state. 
We can, thus, model the human as approximating their preference ordering $\reward^*$ with a reward function $\reward_H: \repspace_H \rightarrow \mathbb{R}$, and their policy mapping $\policy^*$ with $\policy_H(\ac_H \mid \repH) \in [0,1]$. 

The robot can similarly hold representations $\repR$ given by $\repfuncR : \Ob_R^t \rightarrow \repspace_R$. The most general $\repfuncR$ is the identity function, where the robot uses the observations directly, but \secref{sec:reptypes} will also inspect more structured representations. For example, representations can be instantiated as handcrafted feature sets, where the designer distills their prior knowledge by pre-defining a set of representative aspects of the task~\cite{bajcsy2017phri, Ng2000inverse, HadfieldMenell2017InverseRD}, or as neural network embeddings, where the network tries to implicitly extract such prior knowledge from data demonstrating how to do the task~\cite{finn2016gcl, seyed2019smile,xu2019metaIRL}.

\smallskip
\noindent\textbf{Imitation Learning.} Here, the robot's goal is to learn a policy $\policy_R$ that maps from its representation to a distribution over actions $\policy_R(\ac_R \mid \repR)$ telling it how to successfully complete the task. To do so, the robot receives task \textit{demonstrations} from the human and learns to imitate the actions they take at every state~\cite{osa2018algorithmic, torabi2018behavioural}. Let the human demonstration be a state trajectory $\xi = (\st^0, \ldots, \st^\hor)$ of length $\hor$. Importantly, the human and the robot perceive this trajectory differently: the human observes $\xi_H = (\ob_H^0, \ldots, \ob_H^\hor)$ and the robot $\xi_R = (\ob_R^0, \ldots, \ob_R^\hor)$.
Because the demonstrator is assumed to produce trajectories with high reward $\reward_H(\rep_H(\xi_H))$, i.e. be a task expert, the intuition is that directly imitating their actions should result in good behaviour without the need to know the reward.


The issue with this approach is that the human's policy $\policy_H(\ac_H \mid \repH)$ produces actions based on $\repH$, whereas the robot's actions are based on $\repR$. By directly imitating the human, the method, thus, implicitly assumes that $\repH$ is accurately captured by -- or easily recoverable from -- whatever $\repR$ was chosen to be. In other words, it assumes the robot and human's representations of what matters for the task are naturally \textit{aligned}. If this assumption does not hold, the robot might not recover the right policy, and, thus, execute the right actions at the right state.

\smallskip
\noindent\textbf{Reward Learning.} Here, the robot's goal is to recover a parameterized estimate of the human's reward function $\reward_\theta : \repspace_R \rightarrow \mathbb{R}$, from either demonstrations~\cite{ziebart2008maximum,finn2016gcl}, corrections~\cite{bajcsy2017phri}, teleoperation~\cite{javdani2015shared}, comparisons~\cite{christiano2017preferences}, trajectory rankings~\cite{brown2020brex} etc. The intuition here is that the human's input can be interpreted as evidence for their internal reward function $\reward_H$, and the robot can use this evidence to find its own approximation of their reward $\reward_\weight$. Given a learned $\reward_\weight$, the robot can find an optimal policy $\policy_R$ by maximizing the expected total reward $\mathbb{E}_{\policy_R}[\sum_{t=0}^\infty \reward_\theta(\repR)]$.

Similar to imitation, because the human internally evaluates the reward function $\reward_H$ based on $\repH$, their input is also based on $\repH$, whereas the robot interprets it as if it were based on $\repR$. Hence, if the two representations $\repR$ and $\repH$ are \textit{misaligned}, the robot may recover the wrong reward function, and, thus, produce the wrong behaviour when optimizing it~\cite{bobu2020quantifying,fridovich-keil2019confidence}.

\smallskip
\noindent\textbf{The Problem of Misaligned Representations.}
In this paper, we reflect on the traditional assumptions that robot learning are built on and encourage not taking representation alignment for granted:
\begingroup
\addtolength\leftmargini{-0.1in}
\begin{quote}
    \textit{In real-world scenarios, it is unreasonable to assume that robot and human representations will naturally align}.
\end{quote}
\endgroup
We see this in our examples of robot representations $\repR$.
The identity ``representation'' which maps $\obhis_R$ onto itself should, in theory, capture everything in $\repH$ so long as $\obhis_R$ has enough information, but the high-dimensionality of $\Ob_R^t$ makes this representation impractical: learning a reward or policy that is robust across the input space and generalizes across environments would require a massive amount of diverse data -- an expensive ask when working with humans~\cite{Reddy2020SQILIL,fu2018learning}. 
A set of feature functions is lower dimensional, but pre-specifying all features that may matter to the human is unrealistic, inevitably leading to representations $\repR$ that lack aspects in $\repH$~\cite{bobu2020quantifying}. 
Learning neural network embeddings $\repR$ that map from the history $\obhis_R$ while robustly and generalizably covering all $\obhis_R$ (and, thus, $\obhis_H$) requires a lot of highly diverse data, similar to how reward and policy learning on the identity representation would. 
In summary, whether it's insufficient knowledge of what matters for the task or insufficient resources for exhaustively demonstrating the task, the robot's representation will more often than not be misaligned with the human's.

\section{A Formalism for the Representation Alignment Problem in Robotics}
\label{sec:alignment}

How can we mathematically operationalize representation alignment\footnote{For extensions to multiple humans or tasks, see App. \ref{app:extensions}.}?
While it is impossible for the robot and the human to perceive the world the same via $\obhis_R$ and $\obhis_H$, in an ideal world we would want them to \textit{make sense of their observations in a similar way}.
To that end, we formalize the  \textit{representation alignment problem} as the search for a robot representation that is similar to the human's representation. Mathematically, this takes the form of an optimization problem with the following objective:
\small
\begin{equation}
    \repfuncR^* = \arg\max_{\repfuncR} \simfunc(\repfuncR, \repfuncH),
\label{eq:representation_objective}
\end{equation}
\normalsize
where $\simfunc$ is a function that measures the similarity -- or alignment -- between two representation functions.
The key question is: how exactly should we measure representation alignment, i.e. what should $\simfunc$ be? We find the following $\simfunc$ for measuring alignment:
%
\small
\begin{equation}
    \simfunc(\repfuncR, \repfuncH) = -\min_\transmat \sum_{\sthis\in\St^t}\|\transmat^T\repR - \repH \|_2^2 - \lambda\cdot\text{dim}(\repspace_R)\enspace,
    \label{eq:alignment_metric}
\end{equation}
\normalsize
where $\obhis_R$ and $\obhis_H$ correspond to $\sthis$, $\transmat$ is a linear transformation, and $\lambda$ is a trade-off term. We next further explain this notation and why \equref{eq:alignment_metric} best reflects our desiderata from \secref{sec:desiderata}.

\smallskip
\noindent\textbf{D1: Recover the Human's Representation.} To ensure the robot's representation captures \textit{all} relevant task aspects, we intuitively want alignment to be high when the human's representation can be \textit{recovered} from the robot's, no matter the state(s) $\sthis$. Mathematically, we define ``recovery'' as a mapping $\transf : \repspace_R \rightarrow \repspace_H$ from $\repR$ to $\repH$, where $\repH$ is recoverable from $\repR$ if $f(\repR) \approx \repH, \forall \sthis$, where $\obhis_R$ and $\obhis_H$ correspond to $\sthis$. In other words, we can express the recovery error via an $L_2$ distance summed across all state sequences $\sthis$: $ \sum_{\sthis\in\St^t}\|\transf(\repR) - \repH \|_2^2$. In \equref{eq:alignment_metric}, we want representation functions $\rep_R$ that have high alignment $\simfunc$ with $\rep_H$ to have low recovery error, hence we use the negative best distance as a measure of similarity. Note that we chose the $L_2$ distance metric for exposition but other metrics may apply as well. In \secref{sec:survey_metric}, we will survey metrics akin to ours that have been used for comparing representations.

\smallskip
\noindent\textbf{D2: Avoid Spurious Correlations.} We want $\repR$ to not just recover $\repH$, i.e. be sufficient, but also be \textit{minimal} to avoid spurious correlations that reflect irrelevant task aspects. We formalize this with a penalty on the dimensionality of the robot representation function's co-domain $\repspace_R$.
Together, \textbf{D1} and \textbf{D2} describe in \equref{eq:alignment_metric} a measure of representation alignment that rewards small representations that can be mapped close to $\repH$,
%
%
where $\lambda$ is a designer-specified trade-off parameter.

\smallskip
\noindent\textbf{D3: Easily Recover the Human's Representation.} We operationalize the ability to \textit{easily} recover $\repH$ from $\repR$.
Finding an optimal solution to \equref{eq:alignment_metric} via typical optimization methods is intractable given the large space of functions $f$ to search over. In theory, if the human's $\repfuncH$ can be queried by the robot (e.g., by asking for labels), the most straightforward solution collects feedback $\langle \obhis_R, \repfuncH(\obhis_H)\rangle$ from the human
and fits an approximation $\hat\transf(\repR)\approx \repH$, e.g. a neural network. Unfortunately, even if $\repR$ is low-dimensional, fitting an arbitrarily complex $\hat\transf$ that reliably results in high alignment for all states could require a large amount of representative labels, i.e. it would not be \textit{easy} to recover the human's representation.
For this reason, we want ``easy'' recovery to involve a transformation $\transf$ of small complexity.
This condition has been mathematically stated via a multitude of complexity theory arguments (upper bounds based on the Vapnik–Chervonenkis dimensions~\cite{baum1988piecewiseconstant,harvey201relu,bartlett1998piecewisepolynomial,KARPINSKI1997sigmoid} or the Radamacher complexity of the function~\cite{golowich2018radamacher,Bartlett2017SpectrallynormalizedMB}), but recent empirical work argues that linear transformations are a good proxy for small complexity~\cite{Coates2012LearningFR,lai2019contrastive,reed2022separability,alain2017linearprobe}. We thus similarly take $\transf$ to be a linear transformation given by a matrix $\transmat$. 

\smallskip
\noindent\textbf{D4: Explain the Robot's Representation.} Human-aligned representations should be amenable to interpretability and explainability tools. If the human representation is easily recoverable, i.e. the robot can learn a good estimate $\hat\transf$, we get this condition almost for free without encoding it in \equref{eq:alignment_metric}: the robot can communicate its representation to the human by showing examples $\langle \obhis_H, \hat{\transf}(\repR)\rangle$ where observation sequences are labeled with the robot's current ``translation'' of its representation. The last piece we need for explainability is ensuring that $\hat\transf$ is understandable by the human, by, for example, having additional tools that can convert $\hat\transf$ into more human-interpretable interfaces, like language or visualizations. 

\smallskip
\noindent\textbf{Examples of Robot Representations.}
Since solving \equref{eq:representation_objective} is clearly intractable for an arbitrarily large set of functions $\rep_R$, different ways of defining the robot's representation $\repR$ implicitly make different simplifying assumptions.
When $\repfuncR$ is the identity function, the underlying assumption is that there exists some $\transf : \Ob_R^t \rightarrow \repspace_H$ that satisfies \equref{eq:alignment_metric} so long as $\obhis_R$ has enough information to capture $\repH$. 
Unfortunately, because $\transf$ operates on an extremely large space of robot observation histories $\Ob_R^t$, it would have to be complex enough to reliably cover the space, violating \textbf{D3}. This, together with the large dimensionality of the representation space, result in a small alignment value in \equref{eq:alignment_metric}. 
Meanwhile, methods that assume that $\repR$ has some more low-dimensional structure, like the feature sets or embeddings from earlier, could also have small alignment values: 
feature sets might be non-comprehensive, while learned feature embeddings might have not extracted what's truly important to the human, making it, thus, impossible to find an $\transf$ that recovers $\repH$. As we will see in \secref{sec:reptypes}, no representation is naturally human-aligned and every representation type comes with its trade-offs.

\section{A Survey of Robot Representations}

We present our survey of four categories of \textit{learned} robot representations: identity, feature sets, feature embeddings, and graph structures. Table \ref{tab:representations} situates them within our formalism and highlights key tradeoffs. We then additionally compare the representation types by surveying the few works that quantify alignment.

\begin{table*}[h]
\centering
    \captionsetup{justification=centering}
    \caption{Existing representations (and example papers) through the lens of our formalized desiderata.}
    \vspace{-3mm}
    \label{tab:representations}

    {\renewcommand{\arraystretch}{1.5}
    \resizebox{1.0\textwidth}{!}{
    \begin{tabular}{|p{4.9cm}|p{6.4cm}|p{2.4cm}|p{6.8cm}|p{3.0cm}|}
        \hline
        \centering\textbf{Representation Type} & 
        \centering\textbf{D1: Recoverability of} $\repH$ \textbf{from} $\repR$ $\min_\transf \sum_{\sthis\in\St^t}\|\transf(\repR) - \repH \|_2^2$ & 
        \centering\textbf{D2: Minimality} $\text{dim}(\repspace_R)$  & 
        \centering\textbf{D3: Ease of Recovery of} $\repH$ \textbf{from} $\repR$ $\min_\transmat \sum_{\sthis\in\St^t}\|\transmat^T\repR - \repH \|_2^2$ & 
        \textbf{D4: Interpretability} \cr

        \hline

         Identity \newline $\repR = \obhis_R \in \Ob_R^t$  \newline \cite{osa2018algorithmic,torabi2018behavioural,finn2016gcl,christiano2017preferences,finn2017maml,xu2019metaIRL} & 
         \textcolor{ForestGreen}{Contains complete information} & 
         $|\Ob_R^t|$, \textcolor{BrickRed}{Large} & 
         \textcolor{BrickRed}{Difficult in arbitrarily large observation spaces} & 
         \textcolor{BrickRed}{Black box} \\

        \hline

         Feature Set \newline $\repR=\{\rep_R^1(\obhis_R), ..., \rep_R^d(\obhis_R)\}$ \newline \cite{vernaza2012efficient, choi2013bayesian, levine2010feature, ratliff2007boosting, bobu2021ferl,bobu2021perceptual,paxton2021predicting, yuan2021sornet} & 
          \textcolor{YellowOrange}{May lack information but can use misalignment detection methods to learn new features} &
          $d$, \textcolor{YellowOrange}{Grows linearly} &
          \textcolor{ForestGreen}{If complete, easy} \newline
          \textcolor{YellowOrange}{If complete but $d$ large, medium} \newline
          \textcolor{BrickRed}{If incomplete, hard} &
          \textcolor{ForestGreen}{High} \\
          
        \hline

        Feature Embedding (Unsupervised) \newline $\repR=\vec\rep_R(\obhis_R) \in \mathbb{R}^d$ \newline \cite{lesort2018staterepoverview, stooke2021decoupling, schwarzer2021pretraining, anand2019unsupervisedreps, zhang2021invariant, ha2018worldmodels,
        laskin2020curl,
        hafner2020dream} & 
        \textcolor{BrickRed}{May learn wrong disentangled information} &
        $d$, \textcolor{ForestGreen}{Low by design} &
        \textcolor{ForestGreen}{If disentangled information complete, easy} \newline
        \textcolor{BrickRed}{If disentangled information incomplete, hard} &
        \textcolor{YellowOrange}{May be interpretable to the system designer} \\
         
        \hline

        Feature Embedding (Supervised) \newline $\repR=\vec\rep_R(\obhis_R) \in \mathbb{R}^d$ \newline \cite{brown2020brex, tucker2022latentalignment, gleave2018multi, reddy2021pico, hilgard2021mcm, bobu2023SIRL, hristov2019disentangled} & 
        \textcolor{YellowOrange}{More likely to capture relevant information} &
        $d$, \textcolor{ForestGreen}{Low by design} &
        \textcolor{ForestGreen}{If relevant information, easy} \newline
        \textcolor{BrickRed}{If missing relevant information, hard} &
        \textcolor{YellowOrange}{May be interpretable to the system designer} \\
        
        \hline
        
          Graph \newline $\repR = G = \{V, E\}$ \newline \cite{wang2017knowledge, Zareian2020KGSG, daruna2021towardskge, yi2022incremental, chen2020learning, mohseni2019simultaneous, haan2019causal, zhang2021knowledgeextraction} & 
          \textcolor{BrickRed}{May lack information} &
          $|V+E|$, \textcolor{YellowOrange}{$|V|$ linear, $|E|$ quadratic} &
          \textcolor{ForestGreen}{If complete, easy} \newline
          \textcolor{YellowOrange}{If complete but $|V+E|$ large, medium} \newline
          \textcolor{BrickRed}{If incomplete, hard} &
          \textcolor{ForestGreen}{High} \\
        
        \hline
    \end{tabular}
    }}
\end{table*}

\subsection{Robot Representation Types}
\label{sec:reptypes}

\subsubsection{Identity Representation}
As we alluded to in ~\secref{sec:alignment}, an identity representation maps an observation history onto itself, i.e. $\repR = \obhis_R$, with the co-domain of the representation function as the space of observation histories: $\rep_R : \Ob_R^t \mapsto \Ob_R^t$. The methods we review here, thus, don't learn an explicit intermediate representation to capture what matters for the task(s) but instead hope to implicitly extract what's important from human task data.

Because the inputs for reward or policy learning consist of high-dimensional observation histories, e.g. images, we cover approaches based on high-capacity deep learning models. There are now numerous end-to-end methods for learning policies~\cite{osa2018algorithmic,torabi2018behavioural,Reddy2020SQILIL,levine2020offline} or rewards~\cite{finn2016gcl,fu2018learning,wulfmeier2016maxentirl} from demonstrations. These methods perform well with an overparameterized high complexity function but they overfit to the training tasks and suffer from generalization failures due to \textit{distribution shift}~\cite{ross2011reduction}, resulting in arbitrarily erroneous behavior during deployment. Good end-to-end performance across a large test distribution can require thousands of demonstrations for each desired task~\cite{zhang2018deepimitation,Rahmatizadeh2018inexpensive,Rajeswaran2018learning}, which is expensive to obtain in practice. In reward learning, this has been alleviated by introducing other types of reward input like comparisons~\cite{christiano2017preferences}, numeric feedback~\cite{Warnell2018DeepTI}, goal examples~\cite{fu2018variational}, or a combination~\cite{Ibarz2018reward}. These are user friendly alternatives to demonstrations that are amenable to active learning~\cite{reddy2020hypothetical,singh2019activeRL}, further reducing human burden.

Another way to reduce sample complexity is meta-learning~\cite{finn2017maml}, which seeks to learn representations that can be quickly fine-tuned \cite{xu2019metaIRL, yu2019meta, singh2020metaIL, huang2021meta, seyed2019smile}. The idea is to reuse human data from many different tasks; if the training distribution is representative enough, the ``warm-started'' model can adapt to new tasks with little data.
Unfortunately, the human needs to know the test task distribution \textit{a priori}, which brings us back to the specification problem: we now trade hand-crafting features for hand-crafting task distributions.
These models are overparameterized and, thus, are inherently \textit{uninterpretable} and tough to debug in case of failure~\cite{rudin2019stop}.

\smallskip
\noindent\textbf{Takeaway.} In theory, the identity representation contains complete information for recovering the human's representation. However, it is incredibly difficult to use for robust and generalizable robot learning: the dimensionality of the observation space (and of the representation) can be so large that the robot may require an impractically large and diverse set of human task data to reflect every individual, environment, and task it will face. Current trends look at clever ways to cheaply collect human data (e.g. YouTube or VR) or reuse past data from the robot's lifespan. However, there still is no guarantee that this data will be representative of the end user. 

\subsubsection{Feature Sets}
We can instantiate the robot's representation $\repR$ as a set $\{\rep_R^1(\obhis_R), ..., \rep_R^d(\obhis_R)\}$, where each $\rep_R^i(\obhis_R)$ is a different individual dimension of the representation, with $d$ much smaller than $|\Ob_R^t|$. These dimensions represent concrete aspects of the task -- or features, e.g. how far the end effector is from the table, -- which is why we call  $\rep_R^i$ a feature function and the output $\rep_R^i(\obhis_R)$ a feature value. In general, the feature function maps observation histories to a real number indicating how much that feature is expressed in the observations, $\rep_R^i : \Ob_R^t \rightarrow \mathbb{R}$. Hence, under this instantiation, the robot's representation maps from observation histories onto a $d$-dimensional space of real values: $\rep_R : \Ob_R^t \rightarrow \mathbb{R}^d$, where $d$ grows linearly with the number of features.

Handcrafted feature sets have been used widely across policy and reward learning~\cite{abbeel2004apprenticeship,jain2015learning,javdani2015shared,shah2018state}, but exhaustively pre-specifying \textit{everything} a human may care about is impossible~\cite{bobu2018learning}.
To address this, early reward and policy learning methods infer relevant feature functions directly from task demonstrations. \citet{vernaza2012efficient} define the robot's representation as the PCA components of the observations, while other methods specify base feature components for constructing the feature functions as logical conjunctions~\cite{choi2013bayesian, levine2010feature} or regression trees~\cite{ratliff2007boosting}.

Unfortunately, engineering a relevant set of base features can be tedious and incomplete. Moreover, because they use low-capacity learning models for the feature functions, these methods are limited to discrete or low-dimensional observation spaces. Hence, recent approaches propose representing individual feature functions with neural networks~\cite{bobu2021ferl,bobu2022inducing,bobu2021perceptual,paxton2021predicting, yuan2021sornet} and training them with labeled observations~\cite{paxton2021predicting, yuan2021sornet}. ~\citet{paxton2021predicting} learn complex spatial relations mapping from high-dimensional point cloud observations but require large amounts of data, which is impractical for teaching multiple feature functions. One approach reduces this data complexity with a new type of structured input, a feature trace, which yields large amounts of feature value comparisons for training the network with little effort from the human~\cite{bobu2021ferl,bobu2022inducing}. Another approach reduces the burden via bootstrapping, using a small amount of human labels to learn feature functions defined on a lower dimensional transformation of the observation space (object geometries) then using that to label data in a simulator (object point clouds)~\cite{bobu2021perceptual}.

\smallskip
\noindent\textbf{Takeaway.}
Feature sets are helpful for inserting structure in the downstream learning pipeline making it more data efficient, robust, and generalizable~\cite{bobu2022inducing}. However, that added structure is \textit{useful only if correct}: without the right feature sets, robots may misinterpret the users' guidance for the task, execute undesired behavior, or degrade performance \cite{bobu2020quantifying}.
Under-specified feature sets can be handled by detecting misalignment~\cite{bobu2020quantifying} and learning new features, but we need more ways to reduce the human burden for teaching features, like introducing new types of structured input~\cite{bobu2021ferl} or bootstrapping the learning~\cite{bobu2021perceptual}. If, on the other hand, the structure is over-complete, i.e. it contains irrelevant features, it can lead to spurious correlations, which we could prevent via feature subset selection~\cite{cakmak2012questions,bullard2018featureselection, hoai2019refinement}.

\subsubsection{Feature Embeddings}

We review a vast body of work on representations learned as feature embeddings in a neural network. Here, the robot's representation $\repR$ is instantiated as a low-dimensional feature embedding, or vector, $\vec\rep_R(\obhis_R)$, where each dimension is a different neuron in the embedding. The representation function is $\rep_R : \Ob_R^t \rightarrow \mathbb{R}^d$, with $d$ fixed by the designer and much smaller than $|\Ob_R^t|$. While feature set functions also map to $\mathbb{R}^d$, each dimension is learned individually (and is representative of some task aspect), whereas here the embedding is learned jointly (and hopes to capture important task aspects implicitly).
We identify two broad areas in this space: unsupervised methods (also called self-supervised), which use unlabeled data and proxy tasks to learn representations, and supervised methods, which use human supervision at the representation level. We also cover some in-between semi- or weakly-supervised methods.

\smallskip
\noindent\textbf{Unsupervised methods.} 
At the most data-efficient extreme, unsupervised methods try to learn disentangled latent spaces from data collected without any human supervision. Instead of explicitly giving feedback, the human designer hopes to instill their intuition for what is causal for the task by specifying useful \textit{proxy tasks} ~\cite{lesort2018staterepoverview,chen2022empirical,lee2021pebble,yang2021representationsmatter}. 
In robot learning, these proxy tasks range from reconstructing the observation (to ignore irrelevant  aspects)~\cite{finn2015feature,Higgins2017DARLAIZ,ha2018worldmodels,lynch2019play}, to predicting forward dynamics (to capture what constrains movement)~\cite{watter2015control,ha2018worldmodels} or inverse dynamics (to recover actions from observations)~\cite{pathak2018zero}, to enforcing behavioural similarity between observations~\cite{zhang2021invariant,ghosh2019actionablereps,aytar2018playing}, to contrastive losses~\cite{oord2018cpc,laskin2020curl,anand2019unsupervisedreps,stooke2021decoupling}, or some combination~\cite{hafner2020dream,schwarzer2021pretraining}. The proxy task result itself does not matter; rather, these methods are interested in the intermediate representation extracted from training on the proxy tasks. However, because they are purposefully designed to bypass supervision, these representations do not necessarily correspond to human features, rendering explicit alignment challenging. In fact, the cases where the disentangled factors match human concepts are primarily due to spurious correlations~\cite{Louizos2016TheVF}. Lastly, like all learned latent representations, they are difficult to interpret by end users.



\smallskip
\noindent\textbf{Supervised Methods.} At the other extreme, we have human-supervised approaches. Some methods combine the human's task data with self-supervised proxy tasks to pre-train a useful feature embedding~\cite{brown2020brex,tucker2022latentalignment} while others reduce supervision by learning a simpler model that, when trained well, can automatically label large swaths of videos of people doing tasks~\cite{baker2022youtube}.
Multi-task methods pre-train representations from human input for multiple tasks, then fine-tune the reward or policy on top of the embedding at test time~\cite{gleave2018multi,nishi2020fine,yamada2022taskinduced}. Similar to meta-learning, the motivation here is that the robot collects data from many different but related tasks, which it can then leverage to jointly train a shared representation. 
This is more scalable than meta-learning~\cite{mandi2022effectiveness}, but still needs curating a large set of training tasks to cover the test distribution. 

There is a growing body of work directly targeting supervision at the representation level. \textit{Implicit} methods make use of a proxy task for the human to solve and a visual interface that changes based on the robot's current representation~\cite{reddy2021pico,hilgard2021mcm,bobu2023SIRL}. The hope is that if the human can still solve the proxy task well, the underlying representation must contain salient behavioral aspects. If the representation dimensions are interpretable enough, \textit{explicit} learning of representations is also possible by directly labeling examples with the embedding vector values~\cite{hristov2019disentangled, sripathy2022teaching}.
What both these directions have in common is that the representation \textit{is} or \textit{can be} converted into a form that is interpretable to the human, thus opening the possibility of the human providing targeted feedback that is explicitly intended to teach the robot the desired task representation.

\smallskip
\noindent\textbf{Takeaway.} There is a trade-off between the amount of human supervision at the representation level and how human-aligned the learned representations are. ``Supervising'' by coming up with proxy tasks certainly reduces the end user's labeling effort, but may result in misaligned representations. For this reason, the burden falls on the designer to find representative proxy tasks: we now trade hand-crafting features for hand-crafting proxy tasks. On the other hand, direct supervision more explicitly aligns the robot's representation with the human's, but is also more effortful for the user. Future work should explore easier ways to incorporate human input, from active learning to better user interfaces. Overall, these representations tend to be more interpretable than the identity~\cite{ericsson2021transfer}.

\subsubsection{Graphical Structures}


Lastly, we can map observation histories onto a graph $G = \{V, E\}$, i.e. $\repR = G$ with $\rep_R : \Ob_R^t \mapsto \mathcal{G}$.
Many graphical structure instantiations have been used for robot learning and planning, from Knowledge Graphs (KG)~\cite{daruna2021towardskge}, to Directed Graphs~\cite{saxena2014robobrain}, Markov Random Fields~\cite{gunther2018context}, Bayesian Networks (BN)~\cite{Mataric2017robotlifelong}, Hierarchical Task Networks (HTN)~\cite{mohseni2019simultaneous}, etc.
Here, we briefly cover KGs, HTNs, and BNs and discuss their tradeoffs.

\textbf{KGs} are repositories of world knowledge made up of entities, e.g. ``mug'' or ``table'', and relations between them, e.g. ``on top of''. They are particularly useful when robust robot behavior relies on strong task context priors, like interpreting ambiguous user commands~\cite{Zareian2020KGSG,yi2022incremental} or handling partially observable environments~\cite{nyga2018grounding,daruna2021towardskge}. Since their relational structure directly allows for probing the causal effect of a certain representation component on the robot's behavior, they are often leveraged for interpretability~\cite{xian2019KGreasoning,daruna2022knowledgegraph,daruna2021towardskge}. 
Building comprehensive KGs takes considerable human effort, as the entities and relations must be made by the human or learned from large data sets~\cite{wang2017knowledge,liu2022survey,nickel2015review}. Hence, recent methods have instead learned KG \textit{embeddings}, which afford more efficient learning~\cite{wang2017knowledge,niu2021enginekgi}, but at the expense of interpretability.

\textbf{HTNs} are tree-based representations that organize domain knowledge as hierarchies of primitive or compound tasks. 
This technique is advantageous for fast and robust planning~\cite{au2011shop,obst2005soccer,lin1993hierarchical}, but requires well-conceived, well-structured, and comprehensive domain knowledge (primitive tasks and hierarchy) to be successful: if one of the primitives on the optimal plan fails, the representation may not contain enough information to recover~\cite{nejati2006learning,mohseni2015interactive}. Various approaches have tried to learn the primitives themselves~\cite{menache2002subgoals}, the hierarchy given the primitives~\cite{mehta2008automatichierarchy}, or both~\cite{hayes2016autonomousy,chen2020learning,mohseni2019simultaneous,li2014learning}, or has combined HTNs with KGs for extracting the necessary additional information to solve the task when primitives are missing or erroneous ~\cite{nyga2018grounding, daruna2021towardskge}. However, most of these methods rely on a set of hand-specified ``base'' primitives, which are non-trivial to build.

\textbf{BNs} are directed acyclic graphs where the nodes are task variables (e.g. the observation history) and the edges are probabilistic conditional dependencies. 
Many works hand-define a task-specific BN structure and learn the corresponding task probabilities~\cite{song2010constraings,infantes2011learning,Ramachandran2009sarsa,Dearden2005forward,chung2015bayesianIL}. For \textit{learning} the BN structure itself, past work defines nodes as atomic components (e.g., histories of binary observations~\cite{Tetsunari2000adaptation,hutter2008featureDBN,lazkano2007bayesian} or features of observation histories~\cite{montesano2007affordances,lopes2007imitation}), adaptively discretizes the node space~\cite{zhang2021knowledgeextraction,Goldenberg2004tractable} or reduces its dimensionality~\cite{song2011BN,deleu2022structure}, then finds the graph edge structure via heuristic search, but this doesn't scale well to real-world settings. Causal structure learning has looked into constructing the graph based on the causal effect that each variable has on the others~\cite{Constantinou2020saiyan,Pearl2010causality}, even leveraging neural networks to learn causal graphs from data~\cite{li2020causal,haan2019causal,Mahmood2011StructureLO}.

\smallskip
\noindent\textbf{Takeaway.} While graphical structures are more interpretable to users, they require significant human effort to construct and maintain relative to their neural network counterparts. Much like specifying rewards by hand, it is hard to specify all relevant nodes, potentially resulting in under-specification. The more modern embedding-based variants bypass some of that specification burden, but at the cost of data efficiency and interpretability.

\subsection{Measuring Representation Alignment}
\label{sec:survey_metric}

We now survey quantitative metrics of representation alignment in order to compare the above representation types. There is little work that directly addresses the representation alignment problem, so we think of the few we mention here as ``case study'' evaluations further supporting our takeaways in Table \ref{tab:representations}, and we reproduce some of the results in Appendix \ref{app:reproduce}. Each work compares a subset of representation types, but none of them covers graphical structures.

\citet{tucker2022latentalignment} propose a metric akin to our recovery error in \equref{eq:alignment_metric} that measures $L_2$ distance between representations. They compare the identity representation to a supervised feature embedding trained by combining human task data with self-supervised proxy tasks. They find that learning representations as supervised feature embeddings can result in as much as 60\% better alignment than the identity. This is consistent with our survey takeaways in Table \ref{tab:representations}: if the designer chooses the right proxy tasks, the learned embedding is more likely to capture relevant information which helps more easily recover the human's representation.

\citet{bobu2022inducing} use the same $L_2$ distance metric but they compare rewards learned as linear combinations of the representations. This is akin to our recovery error in \equref{eq:alignment_metric} with $\transmat$ as the linear reward weights. They compare the identity with a feature set learned one feature at a time with direct supervision from the human. They find that feature sets result in only a third of the alignment error that the identity does; however, they also find that when the learned features are noisy the alignment error is comparable to the identity. We reproduce these results in Appendix \ref{app:reproduce} and they are consistent with our survey takeaway that good (aligned) structure can be very useful in robot learning, but bad (misaligned) structure hinders. 

Lastly, \citet{bobu2023SIRL} use the $L_2$ distance metric to compare the identity to a VAE-based unsupervised feature embedding and a human-supervised feature embedding. They find that in most cases supervised embeddings are better aligned than either identity or unsupervised embeddings, supporting our takeaway that more direct supervision at the representation level leads to better alignment (which we also confirm in Appendix \ref{app:reproduce}). Additionally, the supervised embedding scores low on alignment when it doesn't receive enough supervision to capture the relevant disentangled information, and unsupervised embeddings are better than the identity if sufficiently disentangled, else they are significantly worse.

Despite the representation alignment literature being sparse, we presented the few works that measure alignment of some form between representations. While not identical to our \equref{eq:alignment_metric}, these metrics still evidence the trends in our survey and in Table \ref{tab:representations}.

\section{Open Challenges}
\subsection{Learning Human-Aligned Representations}
\label{sec:HGR}

\textbf{Designing Human Input for Representation Learning.} 
As the dimensionality of the robot task representation is both smaller than that of the task itself and also shareable between tasks, explicitly targeting human input for learning representations \textit{prior} to learning the downstream task distribution should require less overall supervision.
In light of this, we advocate for exploring methods that allow human users to directly give input informing the robot of the representation itself, rather than task inputs~\cite{bobu2021ferl,hristov2019disentangled,bobu2021perceptual,sripathy2022teaching}. In the survey, we saw several examples of such \textit{representation-specific input} types that are highly informative (and intuitive to understand) about desired representations without being too laborious for a human to give, but many more remain to be explored: comparisons and rankings choosing or ordering behaviors more expressive of a certain feature of the representation, equivalences and improvements finding behaviors similarly or more expressive of the feature, natural language describing the feature, or gaze identifying it. Moreover, we can also explore methods that enable the robot to extract the person's representation by having them solve \textit{representation-specific tasks} -- proxy tasks designed to learn an embedding of what matters from their behavior.
For this to be actionable, we encourage development of new interactive interfaces that afford effective communication of desired human representation labels, such that inexperienced users are able to provide useful input.

\smallskip
\noindent\textbf{Transforming the Representation for Human Input.}
A second complementary approach is to directly design robot representations to resemble those naturally understood by humans.
In some cases, it may be possible for the system designer to transform the full task representation into a form that is more aligned with how humans perceive the task. 
This can happen if the designer has prior knowledge that the class of features the robot needs to learn has a well-studied human representation.
Knowing this, we can instantiate learnable robot representations that are well equipped for soliciting human input of the same form, such as masked image states for visual navigation.
Future work should explore other avenues of leveraging human-comprehensible concepts, such as natural language, for instantiating robot representations \cite{shridhar2022cliport,radford2021learning}. This will be beneficial for not only downstream task learning, but also for forming a shared language by which the robot can effectively communicate to the human what it \textit{thinks} is the correct representation prior to deployment. 

\subsection{Detecting Misalignment}

\noindent\textbf{Robot Detecting Its Own Misalignment.} 
If the robot's representation is misaligned, it may misinterpret the humans' guidance for how to complete the task, execute undesired behaviour, or degrade in overall performance \cite{bobu2020quantifying}. Hence, we wish for the robot to \textit{know when it does not know} the human's representation \textit{before} it starts incorrectly learning how to perform the task.
If misalignment is detected, then the robot can re-learn or expand its existing representation rather than wastefully optimizing an incorrect one. 

There are currently two main approaches for detecting misalignment from robot uncertainty: a Bayesian one based on confidence estimates~\cite{fridovich-keil2019confidence,bobu2018learning,Losey2018IncludingUW, bobu2020quantifying,zurek2021situational} and a deep learning one based on neural ensemble disagreement~\cite{Lakshminarayanan2017ensemble,Sun2021OnCE}.
Unfortunately, building in autonomous strategies for robots to detect their own misalignment remains difficult in many scenarios, especially when there is difficulty in disambiguating between representation misalignment and human noise~\cite{bobu2020quantifying}. This issue often arises from inexperienced users and is inherent to the types of data designers must work with in human-robot interaction scenarios.
A proposed, albeit expensive, method of addressing this challenge is to collect more data to balance out noise, but this solution would not fare well in online learning scenarios where the robot must detect misalignment in real time. We suggest that developing methods for fast, online misalignment detection remains critical for real-world deployment.

\smallskip
\noindent\textbf{Human Detecting Robot Misalignment.}
Future work should also build methods that enable human users to detect when a robot's learned representation is misaligned with their own. While the previous section identified a central challenge in robots needing to disambiguate between human input vs. noise, this challenge would be unnecessary if the tools for identifying a correctly learned representation were instead given to the human themselves, i.e. \textit{a human should know what they want the robot to do}.

In the simplest case, the human detects misalignment by observing behaviour produced by the robot, but such behaviours are rarely informative of the underlying reason for failure~\cite{kwon2018expressing}. Because of this, the field of robot explainability has developed tools that are informative of the causal factors behind a system failure~\cite{das2021semantic,daruna2021continual,daruna2021towardskge,daruna2022knowledgegraph}. Consequently, many methods focus on generating post-hoc explanations for explaining behaviour~\cite{glanois2021survey,bewley2021tripletree,hayes2017improving,sanner2005simultaneous,greydanus2018visualizing}. Unfortunately, in real-world deployments, especially those with the added risk of potential safety hazards, e.g. self-driving cars, users may not have the luxury of being able to observe the consequences of a failed representation \textit{after} the fact. Therefore, a growing body of work has started to build tools for allowing humans to interpret and correct robot representations \textit{prior} to deployment~\cite{reddy2021pico}. We remain hopeful that this is a promising direction, and suggest that building in mechanisms for humans to explicitly correct representations should be an integral part of the learning process.

\subsection{Evolving a Shared Representation}

It is also possible for the robot to hold a more complete representation that it wishes to communicate to the human, i.e. teach the human new aspects of the task that they were not aware of before. This may occur in cases of partial observability, where the robot's $\obhis_R$ contains information valuable to solving the task that are not captured by the human's $\obhis_H$ (say, the robot can see a useful tool that the human cannot), or incomplete knowledge, where the robot has knowledge of how to leverage an aspect shared by $\obhis_H$ and $\obhis_R$ that the human does not (say, the robot knows how to use a tool in a way that the human does not). One way for the robot to communicate this information is to show the human examples $\langle \obhis_H, \hat{\transf}(\repR)\rangle$ where observations are labeled with the robot's estimate of the representation transformation function. We can also envision a situation where neither the robot nor the human individually hold a complete representation, and must jointly communicate missing aspects of the desired representation.
By alternating between the direct (robot learning about the human's representation) and the reverse (robot teaching the human about its representation) channels of communication, we can enable reaching a mutual representation that is most informative to completing the task.

\section{Takeaways}

In this work, we proposed a formal lens for viewing the burgeoning field of \textit{representation alignment} in robot learning. We mathematically defined the problem, identified four key desiderata, situated current methods within this formalism, and highlighted their key tradeoffs. Our paper is untraditional in that it is a part-survey, part-formalism retrospective that we hope sheds light on the current gaps present in representations for robot learning and opens the door to exploring future directions and challenges in HRI.

A limitation of our retrospective is that we do not offer a practical solution for \equref{eq:alignment_metric}. Despite this, we believe there is still tremendous value in explicitly formalizing representation alignment beyond simply reviewing literature. 
First, explicitly distilling the four identified desiderata into a unified \equref{eq:alignment_metric} enables researchers to bring broader ideas from the general learning literature into human-robot interaction in a principled way, i.e. we can now take inspiration from general methods to tackle representation alignment, and, thus, then solve HRI-specific problems. For instance, take desideratum 2’s mandate that the desired robot representation be “minimal”. Translating this notion into the dimensionality reduction term of \equref{eq:alignment_metric} enables us to see a direct connection between the rich literature for representation compression, e.g. information bottleneck methods, and its applicability for learning human-aligned representations in HRI. Such a solution from a broader learning principle applied to a HRI-specific problem may have seemed previously unrelated, but can now be found through the lens of \equref{eq:alignment_metric}. We believe there are many other similar opportunities for connecting general machine learning insights to representation alignment applied to HRI.

Moreover, our proposed formalism allows HRI researchers to identify gaps in current methods (including the ones in Table \ref{tab:representations}) and provide directions for future work. Existing work serves as case studies, fulfilling some desiderata but falling short on others.

Lastly, defining representation alignment as the complex optimization problem in \equref{eq:alignment_metric} allows us to assess future methods based on how well they approximate solutions to the full problem. We hope future work will seek novel approximations to \equref{eq:alignment_metric} to explicitly and rigorously tackle this important challenge.

\bibliographystyle{ACM-Reference-Format}
\balance
\bibliography{0-main}

\appendix

\newpage
\appendix
\section{Appendix}

\subsection{Extensions to Formalism in Section \ref{sec:alignment}}
\label{app:extensions}

\noindent\textbf{Extension to Multiple Tasks.} In Sec. \ref{sec:alignment}, we considered the single task setting, where the robot's goal is to successfully perform one desired task. However, our formalism can be extended to account for multiple tasks. First, when the person wants to train the robot to correctly perform multiple tasks, the observation space $\Ob_R$ may be different for each task. In practice, these observation spaces are oftentimes the same or similar (e.g. multiple robot manipulation tasks can all still use images of the same tabletop as observations, although the observation distribution may differ if different objects are used). We can account for differing spaces by choosing the overall observation space $\Ob_R$ to be the union of all individual $N$ task observation spaces $\Ob_{R_i}$: $\Ob_R = \Ob_{R_1} \bigcup  ... \bigcup \Ob_{R_N}$. 
Additionally, in multi-task settings, the human representation $\repH$ will reflect aspects of the task \textit{distribution} that matter to them, rather than of a single task. As a result, the robot's representation learning strategy should reflect this, as we will see in the survey portion of the paper. 

\smallskip
\noindent\textbf{Extension to Multiple Humans.} Aligning the robot's representation to multiple humans requires acknowledging that each human may operate under a different observation space $\Ob_H$ or representation $\repH$. First, we could modify our formalism for differing spaces similarly to how we did in the multi-task setting, by choosing the overall observation space $\Ob_H$ to be the union of all individual $M$ human observation spaces $\Ob_{H_i}$: $\Ob_H = \Ob_{H_1} \bigcup  ... \bigcup \Ob_{H_M}$. Second, in such multi-agent settings, the robot could attempt to align its representation to a unified $\repH = \phi_{H_1}(\obhis_H) \bigcup ... \bigcup \rep_{H_M}(\obhis_H)$, individually to each $\rep_{H_i}(\obhis_H)$, or a combination of the two strategies where the unified representation is then specialized to each individual human's representation.

\subsection{Reproducing Results in Section \ref{sec:survey_metric}}
\label{app:reproduce}

The works we highlighted in Section \ref{sec:survey_metric} serve as ``case study'' evaluations to further support our Table 1 takeaways. Here, we additionally reproduce the results in ~\cite{bobu2022inducing} and ~\cite{bobu2023SIRL}, and replot them using the metric in Eq. \ref{eq:alignment_metric} with a fixed $\lambda=0.0001$ and ground truth $\phi_H$.

\begin{figure}[H]
\centering
\includegraphics[width=0.47\textwidth]{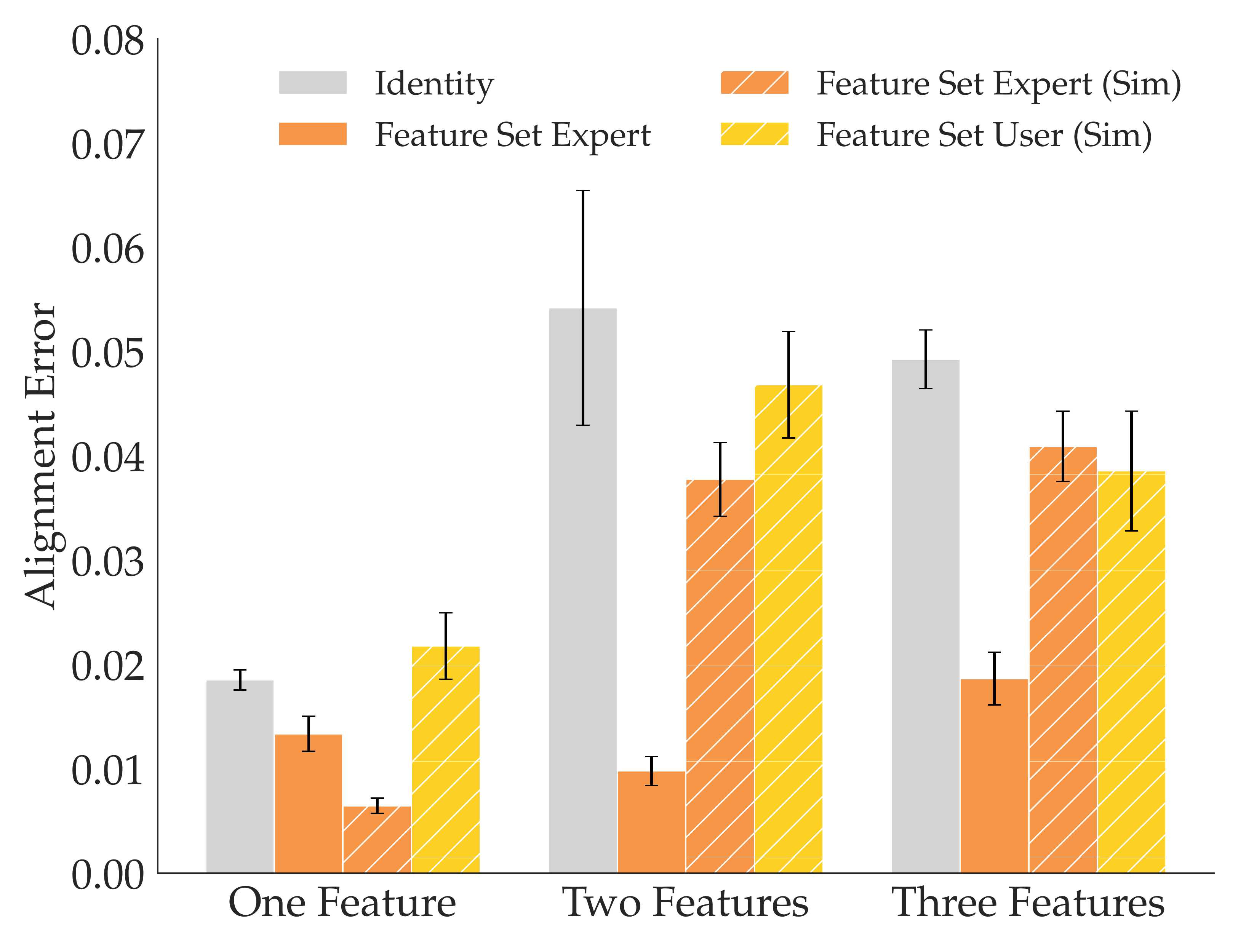}
\caption{Reproducing alignment comparison in ~\cite{bobu2022inducing}. Good feature sets (orange solid) exhibit more alignment than the identity representation (gray), but the noisier the features are (hashed orange and yellow), the less aligned the learned representations are.}
 \label{fig:FERL_reproducing}
\end{figure}

First, in Figure \ref{fig:FERL_reproducing} we compare the alignment error using data from the original ~\citet{bobu2022inducing} Fig. 14 and 17. We compare the \textit{Identity} representation with a feature set trained with expert human data on a real robot manipulation task (\textit{Feature Set Expert}), where the ground truth $\phi_H$ is comprised of \textit{One}, \textit{Two} or \textit{Three} features. We also compare to the ``noisier'' equivalents of the feature set with data in a simulator (\textit{Feature Set Expert (Sim)}) as well as from novice users in a study (\textit{Feature Set User (Sim)}). We find that good feature sets exhibit more alignment than the identity, but when many learned features are noisy (due to losing fidelity in the simulator or to novice user data) the alignment gap decreases. This is consistent with our Table \ref{tab:representations} takeaway that good (aligned) structure can be very useful in robot learning, but bad (misaligned) structure hinders. 

\begin{figure}[H]
\centering
\includegraphics[width=0.35\textwidth]{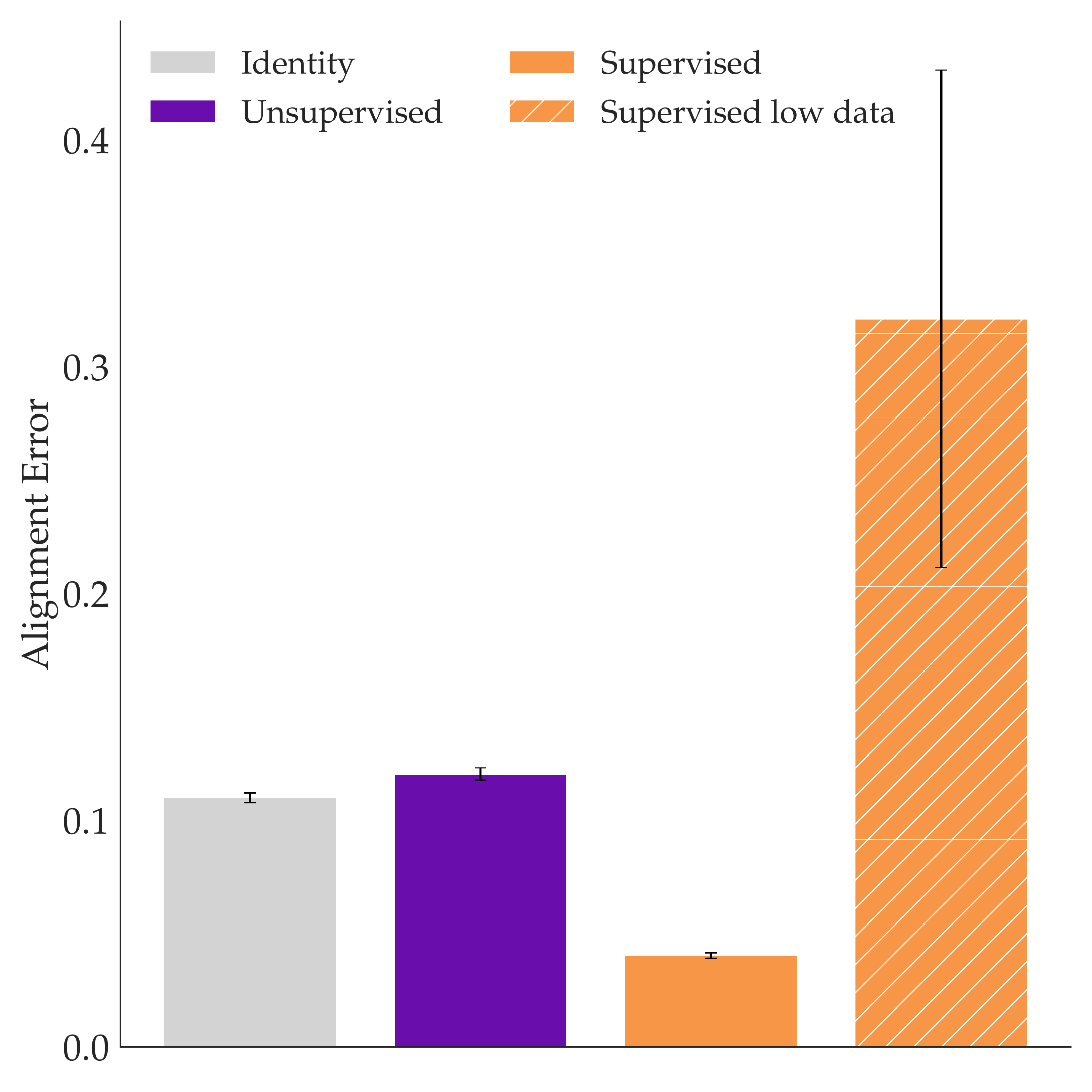}
\caption{Reproducing alignment comparison in ~\cite{bobu2023SIRL}. Good supervised embeddings (orange solid) exhibit more alignment than the identity (gray) or unsupervised embeddings (purple). However, when the embeddings don't have enough supervision (orange hashed), they learn the wrong structure which is detrimental to alignment.}
\label{fig:SIRL_reproducing}
\end{figure}

Second, in Figure \ref{fig:SIRL_reproducing} we compare the alignment error using data from the original ~\citet{bobu2023SIRL} Fig. 4. We compare \textit{Identity} with a VAE-based unsupervised feature embedding (\textit{Unsupervised}) and a human-supervised feature embedding (\textit{Supervised}). We also provide results for an embedding supervised with not enough data (\textit{Supervised low data}). The ground truth $\phi_H$ is comprised of 4 features in robot manipulation tasks. We find that good supervised feature embeddings are better aligned than either identity or unsupervised feature embeddings. However, when they don’t have enough human data, supervised embeddings score low on alignment. 
In this environment, unsupervised embeddings are on par with or slightly worse than the identity. The explanation in the original paper is that the 7DoF robot manipulation environment is complex enough that the VAE can't learn the correct disentangled information. However, their paper additionally presents results in a simpler gridworld environment where unsupervised embeddings perform better as they have an easier time disentangling the right factors of variation. These results are consistent with several takeaways in Table 1.

\end{document}